\documentclass[11pt]{article}
\usepackage{bbm}
\usepackage[]{mdframed}
\usepackage{xcolor}
\usepackage{adjustbox}
\usepackage{comment}
\usepackage{multirow}

\usepackage{physics}
\usepackage{acl}
\usepackage{array}
\usepackage{color,soul}
\usepackage{placeins}
\usepackage{amssymb}
\usepackage{booktabs, multirow, array, makecell, caption}

\usepackage{xcolor}
\usepackage{times}
\definecolor{aqua}{rgb}{0.0, 1.0, 1.0}
\usepackage{latexsym}
\usepackage[ruled,linesnumbered]{algorithm2e}
\RestyleAlgo{ruled}
\usepackage{tabularx}
\usepackage{hyperref}
\usepackage{pifont}
\usepackage{verbatim}
\usepackage{soul}
\newcommand{\hlc}[2][yellow]{{%
    \colorlet{foo}{#1}%
    \sethlcolor{foo}\hl{#2}}%
}

\usepackage{subcaption}
\usepackage{mwe}

\usepackage{mathtools}
\usepackage{lipsum}
\usepackage{float}
\newcolumntype{b}{X}
\newcolumntype{s}{>{\hsize=.5\hsize}X}

\usepackage[T1]{fontenc}
\usepackage{microtype}

\title{Chain-of-Dictionary Prompting Elicits Translation \\in Large Language Models}

\author{
    Hongyuan Lu$^\heartsuit$\Thanks{\hspace{1mm}Equal Contribution.}, Haoran Yang$^{\heartsuit*}$, Haoyang Huang$^\spadesuit$\\
    \textbf{ Dongdong Zhang$^\spadesuit$,Wai Lam$^\heartsuit$, Furu Wei$^\spadesuit$}\\
    $^\heartsuit$The Chinese University of Hong Kong\\
    $^\spadesuit$Microsoft Corporation\\
    \{hylu,hryang,wlam\}@se.cuhk.edu.hk\\
    \{haohua,dozhang,fuwei\}@microsoft.com
}

\begin{document}
\maketitle
\begin{abstract}
Large language models (LLMs) have shown surprisingly good performance in multilingual neural machine translation (MNMT) even if not being trained explicitly for translation. Yet, they still struggle with translating low-resource languages. As supported by our experiments, a bilingual dictionary between the source and the target language could help. Motivated by the fact that multilingual training effectively improves cross-lingual performance, we show that a chained multilingual dictionary with words expressed in more languages can provide more information to better enhance the LLM translation. To this end, we present a novel framework, \textsc{CoD}, Chain-of-Dictionary Prompting, which augments LLMs with prior knowledge with the chains of multilingual dictionaries for a subset of input words to elicit translation abilities for LLMs. Experiments indicate that ChatGPT and InstructGPT still have room for improvement in translating many language pairs. And \textsc{CoD} elicits large gains by up to 13x chrF++ points for MNMT (3.08 to 42.63 for English to Serbian written in Cyrillic script) on FLORES-200 full devtest set. We demonstrate the importance of chaining the multilingual dictionaries, as well as the superiority of \textsc{CoD} to few-shot in-context learning for low-resource languages. Using \textsc{CoD} helps ChatGPT to obviously surpass the SOTA translator NLLB 3.3B.\footnote{Code and resources available at \url{https://github.com/HongyuanLuke/Chain-of-Dictionary}.}
\end{abstract}
\section{Introduction}
Large language models (LLMs) possess the ability to carry out high-quality machine translation tasks without specific training, as observed in previous studies \citep{NEURIPS2020_1457c0d6,2021arXiv211210668L,2022arXiv221105100W,2022arXiv220501068Z,2023arXiv230304048W}. The models can be prompted to do so by requesting them to complete a prompt, such as “Translate the following sentence to English from French:” followed by an input sentence written in French. However, despite their training on extensive datasets, these models may encounter difficulties in correctly translating rare words that frequently occur in low-resource situations.
\par
\begin{figure*}[t!]
\begin{center}
\vspace{0mm}
\centerline{
%\hspace{-10mm}
\includegraphics[width=15cm]{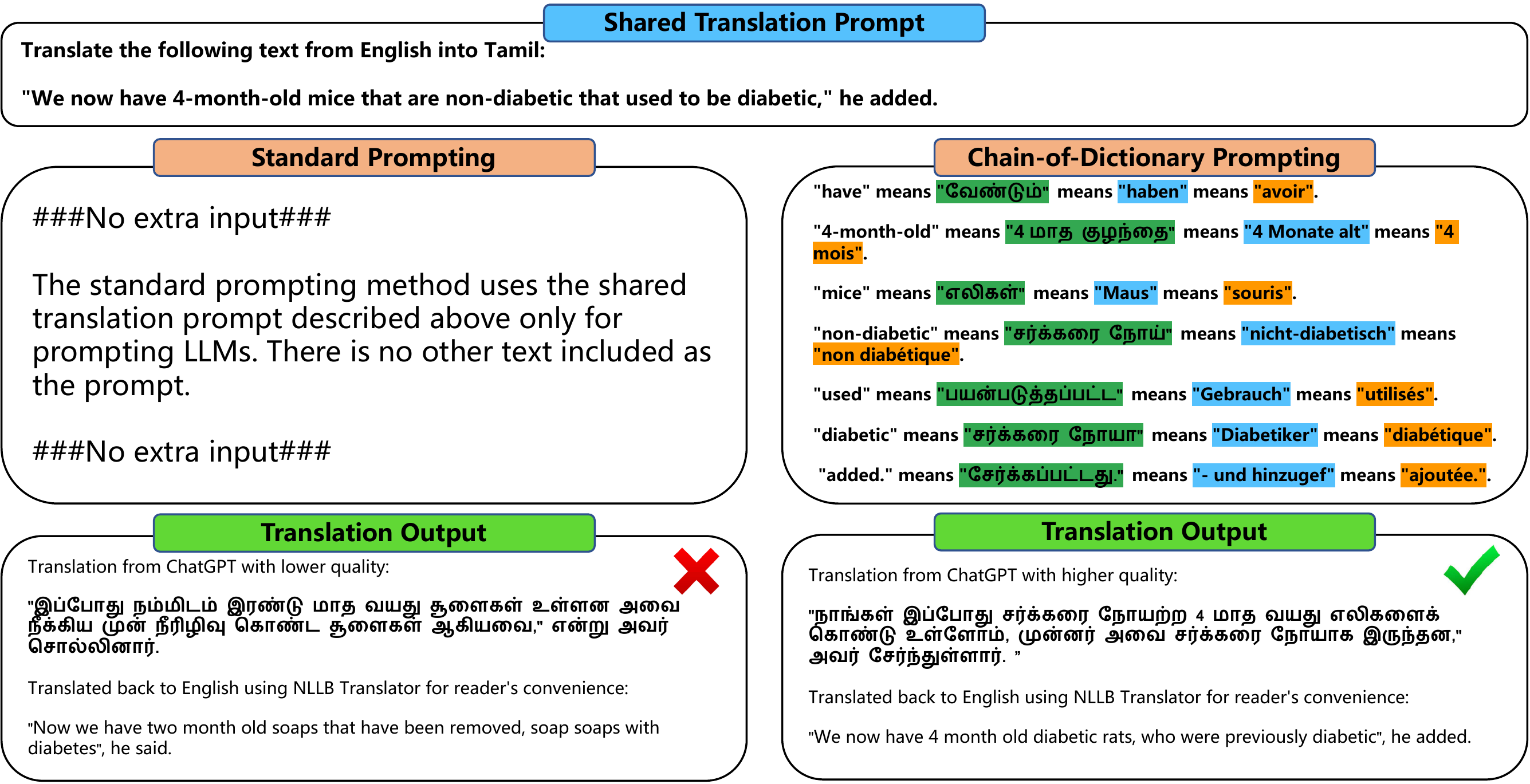}}
    \caption{An illustration for \textsc{CoD} for English to Tamil translation. \textsc{CoD} consists of two sections: the standard translation prompt (the upper box) and the chained multilingual dictionaries. We highlight by languages the chained dictionary part for \textsc{CoD}, containing the words and their translations in different languages. \textsc{CoD} outperforms standard prompting in this example, and other methods such as the conventional Chain-of-Thought have been shown as less effective for MT \citep{2023arXiv230313780P}. We bold the text for the actual inputs/outputs. Other non-bolded texts are placed for the explanation to the readers.}
    \label{fig:cod}
\end{center}
\vspace{-5mm}
\end{figure*}
\par 
Motivated by such a lexical-level problem, we seek how to incorporate dictionaries for improving MNMT. Further, motivated by the fact that multilingual training effectively improves cross-lingual performance \citep{liu-etal-2020-multilingual-denoising,2022arXiv221207752L}, we use multilingual dictionaries to enhance the translation performance of LLM prompting.
\par
To this end, we leverage the multilingual dictionaries as the prior knowledge, and we describe a method to prompt LLMs with hints that indicate a set of possible chained multilingual translations for specific words in the input. This method involves adding a string such as  “`limit' means `Grenze' means `çäk'.”\ to the start of the standard machine translation prompt as lexicon hints for MT. This approach is motivated by the fact that supervised machine translation models have effectively used dictionaries to enhance translation \citep{2016arXiv161007272Z,arthur-etal-2016-incorporating,zheng-etal-2021-non-parametric}. We also propose the method as a chain of dictionary in the light of Chain-of-Thought (CoT) reasoning \citep{2022arXiv220111903W} that represents the reasoning procedure as intermediate thinking steps.  In our case, we show how to incorporate multilingual knowledge in a zero-shot manner by chaining the translations of words across various languages to improve LLM's MNMT capabilities. This allows us to specify the task in the prompt and provide background knowledge that is useful in completing the task of machine translation, without placing any strict constraints on how the model employs this knowledge, as demonstrated in Figure \ref{fig:cod}.
\par
We conducted extensive experiments with the novel framework we propose, namely \textsc{CoD} (\textbf{C}hain-\textbf{o}f-\textbf{D}ictionary Prompting for Machine Translation), which achieved notable improvements in low-resource translation on FLORES-200 benchmarks \citep{nllb2022} between English to almost all the other languages, using various language models. To gain a better understanding of \textsc{CoD}'s capabilities, we analyzed and examined the model's behaviour by comparing it to both settings that incorporate bilingual dictionaries as well as separating the word mappings instead of chaining the multilingual dictionaries. \textsc{CoD} achieves the best empirical performance, which demonstrates its necessity in chaining the multilingual dictionary. Also, our experiments demonstrate that \textsc{CoD} achieves better performance than the standard few-shot demonstrations for low-resource languages. We speculate that the retrieved few-shot demonstrations are not relevant to the target translation, and therefore not particularly useful for low-resource translations.
Our main contributions are three-fold:

\begin{itemize}
\setlength\itemsep{0em}
    \item This paper proposes a novel framework called \textsc{CoD} (\textbf{C}hain-\textbf{o}f-\textbf{D}ictionary Prompting for Machine Translation) which adds chains of multilingual dictionaries to prompt LLMs that substantially improve machine translation.
    \item We conduct experiments on FLORES-200 for all translation directions between English and other languages. We observe that ChatGPT and InstructGPT still have room for improvement in translating many language pairs. We found that \textsc{CoD} can improve ChatGPT on a large portion of the languages, and can elicit translation in some languages that ChatGPT almost completely fails in translating.
    \item We observe that \textsc{CoD} can also be favourable to few-shot demonstrations, and \textsc{CoD} on ChatGPT can even surpass the SOTA translator NLLB 3.3B. We also verify that it is possible to save computation by truncating stopwords from the dictionary.
\end{itemize}
\section{Chain-of-Dictionary Prompting for Neural Machine Translation}
\label{se3}

Large language models show their promising translation performance when sufficiently pre-trained \citep{2022arXiv221207752L, 2023arXiv230304048W}. However, this is frequently not the case, especially for these low-resource languages. There are thousands of languages around the world, and current research on MT has scaled to at least 200 \citep{nllb2022}. It is an important research topic to explore the capabilities of LLMs to cover as many languages as possible. Despite the importance of covering low-resource languages in LLMs, we will report in this paper that the latest LLMs are still far from satisfying in covering these low-resource languages from FLORES-200 \citep{nllb2022}. 
\par 
We propose a novel framework called \textsc{CoD} (\textbf{C}hain-\textbf{o}f-\textbf{D}ictionary Prompting) to address these difficulties by chaining multilingual dictionary knowledge into prompting-based machine translation. Compared to in-context learning that uses few-shot demonstrations to prompt the LLMs, dictionaries are comparatively easier to store and acquire than the demonstrations, particularly for low-resource languages \citep{2016arXiv161007272Z,arthur-etal-2016-incorporating,10.1145/3377713.3377801,2023arXiv230207856G}. This makes \textsc{CoD} an attractive external resource for MT with LLMs.
\par
Our novel approach, \textsc{CoD}, utilizes prompting-based translation and integrates chained multilingual dictionary information as prior knowledge directly into the prompt. When presented with a source sentence, we search for the multilingual dictionary entries for a subset of the words: before making the conventional translation request to LLMs, we append additional textual inputs to the prompt that outline possible chained multilingual translations for those specific words.
\par
Therefore, the prompts for each sentence consist of two parts, as illustrated in Figure \ref{fig:cod}:
\begin{mdframed}
(1) the translation prompt: 
\textit{“Translate the following
text from <source-language> into <target-language>: <source-sentence>”}.
\\
(2) the chained multilingual dictionaries: 
\textit{“<word X in source-language> means <word X in target-language> means <word X in auxiliary-language 1> means  <word X in auxiliary-language 2>.”}; 
\end{mdframed}
\par
We do not include few-shot in-context learning in our methodology as we inspected that it is usually hard to retrieve relevant demonstrations for low-resource languages, which yields limited improvements. In the remaining sections, we will report relevant experimental results which indicate that few-shot demonstrations are less favourable to our methods for low-resource translations.

We also found that using non-chained decomposed multilingual dictionaries instead of \textsc{CoD} degrades the results: 
\par 
\textit{“<word X in source-language> means <word X in target-language>. <word X in source-language> means <word X in auxiliary-language 1>. <word X in source-language> means <word X in auxiliary-language 2>.”}\footnote{We also attempted using different linking words such as “-” and “translates to” instead of “means”, where on-par performance is spotted. Also, note that keeping the dictionary word order to their order of appearance in the source sentence is important. Shuffling the word order can degrade the results.}
\par 
We evaluate Machine Translation performance for all available languages using the LLM which we subsequently enhance with \textsc{CoD}. We then employ top languages that report the highest evaluation scores as our auxiliary languages to construct our multilingual dictionaries.
\paragraph{Multilingual Dictionary} We propose to use the prompt “\textit{Extract the words from the following texts: <input-sentence>}” to extract the keywords from the source language with LLMs such as ChatGPT. We then translate the extracted words into different languages with off-the-shelf MT models such as NLLB to create the dictionaries for \textsc{CoD}. During inference, the matched keywords and their translations are extracted from the dictionary to be appended to the translation prompt.
\par
We use French (fra\_Latn), German (deu\_Latn), and Portuguese (por\_Latn), three high-resource languages that our LLM performs well on, as our auxiliary languages for multilingual dictionaries. This means that we have a chain of 5 languages in the prompt, including the three auxiliary languages mentioned above and the source and the target language. We leave the exploration of further chaining to future work.
\section{Experimental Setup}
\subsection{Baselines}
We experiment with ChatGPT, a multilingual large language model that has shown strong abilities for the task of machine translation \citep{2023arXiv230304048W}. At the time of writing, this LLM was widely popular. We experiment with ChatGPT to test \textsc{CoD}. We also conduct experiments on InstructGPT with the version of text-davinci-003 as well as BLOOM-7b \citep{2022arXiv221105100W}:
\begin{itemize}
\setlength\itemsep{0em}
    \item \textbf{GPT-3.5-TURBO} We use a ChatGPT model GPT-3.5-TURBO accessed via the official API through Python. All paired results are run within a week for fair comparison.
    \item \textbf{TEXT-DAVINCI-003} This is one of the InstructGPT models accessed via the official API provided by OpenAI through Python.
    \item \textbf{BLOOM} BLOOM \citep{2022arXiv221105100W} is an open-sourced LLM trained in 46 natural languages. We use its 7B version as our baseline without any further tuning in this paper.
    \item \textbf{NLLB} NLLB \citep{nllb2022} is an open-sourced SOTA translator. We use its 3.3B version as our baseline.
\end{itemize}
Based on the different versions of GPT models, we use the following prompting methods as the baselines to be compared:
\begin{itemize}
\setlength\itemsep{0em}
    \item \textbf{Monolingual Dictionary}: This is a baseline that uses a monolingual dictionary that contains the words from the target language only.
    \item \textbf{Bilingual Dictionary}: This is a baseline that uses a bilingual dictionary for prompting large language models on the task of machine translation \citep{2016arXiv161007272Z,arthur-etal-2016-incorporating,10.1145/3377713.3377801,2023arXiv230207856G}. It replaces the multilingual dictionaries in blue from Figure \ref{fig:cod} with a bilingual dictionary built with the source language and the target language for the task of MT.
    \item \textbf{Decomposed Dictionary}: This is a baseline that removes the chaining of the dictionary and replaces the chained multilingual dictionaries in blue from Figure \ref{fig:cod} with decomposed multilingual dictionaries. Refer to Section \ref{se3} for more details of this baseline model.
    \item \textbf{Few-shot Demonstration}: This is a baseline that does not use any dictionary. Instead, it retrieves from FLORES-200 devtest the top one/three translation pairs that are semantically similar to the current input translation, measured by BertScore \citep{bert-score} using the English sentences. 
\end{itemize}
\subsection{Datasets and Evaluation Metrics}
For our evaluations on the task of machine translation for various languages including many low-resource languages, we use the dev-test division from FLORES-200 benchmarks \citep{nllb2022}, There are 1,012 sentences included in the dataset, which were extracted from English Wikipedia covering a variety of topics and domains. These sentences have been manually curated by professional translators into about 200 languages. 
\par
We report on all the languages in FLORES-200 for both directions from English and into English. 
\par 
For the evaluation metrics, we report the chrF++ \citep{popovic-2015-chrf} and the BLEU \citep{papineni-etal-2002-bleu} evaluations provided by the sacreBLEU repository.\footnote{https://github.com/mjpost/sacrebleu} We use the model [eamt22-cometinho-da]\footnote{\url{https://github.com/Unbabel/COMET}} for generating the COMET scores \citep{rei-etal-2020-comet}.
\subsection{Dictionaries}
To create the offline dictionaries used in our experiments, we first  use the prompt “\textit{Extract the words from the following texts: <input-sentence>}” to extract the keywords from the source language with LLMs such as ChatGPT. We then use the NLLB translator\footnote{https://huggingface.co/spaces/Narrativaai/NLLB-Translator} to translate the monolingual English corpus from FLORES-200 into the remaining languages as our dictionaries. We excluded three languages which are not supported by the NLLB translator from our experiments. We use an off-the-shelf stopwords list for experiments on truncating stopwords to save computations with \textsc{CoD}.\footnote{https://gist.github.com/sebleier/554280}
\par 
We use the English corpora from FLORES-200 to create our dictionary in this paper. For experiments on translating into English, we remove the English reference words from the dictionary to ensure there is no information leakage.
\subsection{Polysemy} With NLLB 3.3B, we translated the words into rare words with multiple attempts and translated back them into English. We then asked ChatGPT whether the translated-back version had the equivalent meaning to the original English. The process was done repeatedly until GPT reported that they were the same or the max tries (3 times) had been hit. In this manner, 71\% of the words are successfully translated without hitting the max tries. For those failed translations, we exclude them from the dictionaries used by the bilingual chain or \textsc{CoD}.
\subsection{Prompting Design}
This section outlines the prompt design we opted for in creating the green text depicted in Figure \ref{fig:cod}.
\par
Prior work compared various prompts for machine translation on LLM \cite{2023arXiv230304048W}, and they have found similar performance of different prompts reported on a limited number of languages. They have opted for a basic prompt \textit{“Translate the following text into <target-language>: <source-sentence>”} as their best prompt. In contrast, our preliminary experiments show that removing the source language name can hurt the performance of translation. Therefore, we opted for \textit{“Translate the following text from <source-language> into <target-language>: <source-sentence>”}.
\par
Our preliminary experiments show that missing the keyword `Tradition Script' for Chinese prompts the model to keep generating Simplified Chinese. Therefore, we specify the language script in our prompt when the languages can be written in different scripts and should be differentiated. For example, we write “Achinese with Arabic script” for the language “ace\_Arab”.
\begin{figure*}[t!]
\begin{center}
\vspace{0mm}
\centerline{
%\hspace{-10mm}
\includegraphics[width=15cm]{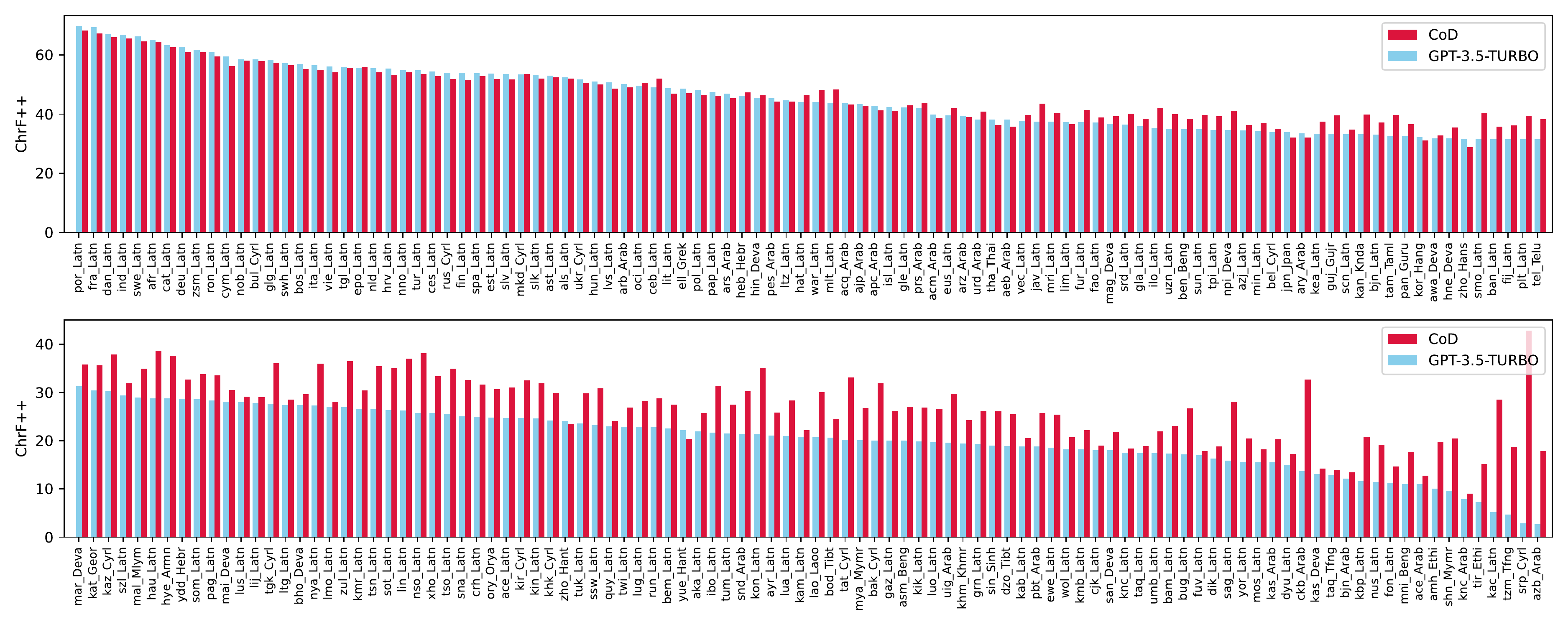}}
\caption{An illustrated comparison of 200 languages from English into the languages between the baseline ChatGPT (GPT-3.5-TURBO) and \textsc{CoD}. We sorted the language scores in chrF++ for ChatGPT in descending order, and we split the whole figure into two parts for clarity. We present the first half in the upper figure, and we present the second half in the bottom figure. \textsc{CoD} is effective for many languages, especially for low-resource ones.}
\label{cod_f}
\end{center}
\vspace{-3mm}
\end{figure*}
\begin{table*}
\centering
    \setlength\tabcolsep{3.5pt}
    \setlength\extrarowheight{0pt}
\begin{tabular}{c|cccc|ccc}
\hline
\noalign{\vskip 1mm}  
\textbf{Direction} & \textbf{\# improved} & \textbf{> 5 points} & \textbf{> 10 points} & \textbf{> 20 points} & \textbf{\# degraded} & \textbf{> 5 points} & \textbf{> 20 points}\\
\noalign{\vskip 1mm}  
\hline
\noalign{\vskip 1mm} 
X-En  & 200/200  & 200/200 & 200/200 & 197/200 & 0/200 & 0/0 & 0/0\\
En-X  & 135/200  & 71/135 & 13/135 & 2/135 & 65/200 & 2/65 & 0/65\\

\noalign{\vskip 1mm}  
\hline
\end{tabular}
\caption{\label{summ}
Statistics of the changes in chrF++ with \textsc{CoD} on GPT-3.5-TURBO with 200 languages. 83.75\% of the directions (335 out of 400) are improved. The advantage of \textsc{CoD} clearly outweighs the disadvantage.  
}
\end{table*}
 
\section{Results and Analysis}
\subsection{En-X Results}
\paragraph{En-X: ChatGPT} We firstly compare ChatGPT (GPT-3.5-TURBO) with the normal prompt in chrF++ on FLORES-200 with \textsc{CoD}. We plot the results in Figure \ref{cod_f} for better clarity. In Figure \ref{cod_f}, we sort the chrF++ scores from ChatGPT in descending order, and we split the whole results into two figures. The upper figure represents the first half, and the bottom figure represents the second half. It can be observed in the bottom figure that ChatGPT does not handle the translation perfectly and it reports a score under 30 points in chrF++ for around 100 out of the 200 languages. The results indicate that \textsc{CoD} brings clear improvements. For space reasons, we leave Table \ref{enx_results} in the Appendix to present the detailed results for translating from English into the remaining languages. Table \ref{bleu} in the Appendix also reports the detailed BLEU evaluations. Those results also indicate strong improvements with \textsc{CoD}. We speculate there are two reasons for improvement with \textsc{CoD}. Firstly, putting the desired translation target lexical shrinks the translation space and eases the translation.  Secondly, using auxiliary languages in the chain
gives better cross-lingual cues when there is no direct mapping between source and target lexical.
\par
\paragraph{En-X: Languages Improved on ChatGPT} Table \ref{summ} reports that more than 67\% (135 out of 200) of the languages can be improved by \textsc{CoD}. For those languages that can be improved by \textsc{CoD}, more than 50\% (71 out of 135) is improved by at least 5 points in chrF++. 13 languages can be improved by at least 10 points in chrF++ and 2 languages can be improved by at least 20 points in chrF++. We also observe quite strong results with \textsc{CoD} that bring 13x improvement (3.08 to 42.63) when translating from English into Serbian written in Cyrillic script. This leads to the conclusion that \textsc{CoD} gives promising results with good improvements in most languages and excellent improvements in several languages. \textsc{CoD} can even elicit translation in some languages that ChatGPT almost completely fails in translating, which is quite promising.
\paragraph{En-X: Languages Not Improved on ChatGPT} As in Table \ref{summ}, some languages are not benefited from \textsc{CoD}. We observe there are no languages with more than 20 points of decrease in chrF++ with  \textsc{CoD}, and there are only 2 languages with more than 5 points of decrease in chrF++ with \textsc{CoD}. Compared to the languages with improvements reported above, the advantages of using \textsc{CoD} clearly outweigh the disadvantages when used indistinguishably regardless of the languages.
\par 
\paragraph{En-X: Languages Selection} Though one could use \textsc{CoD} regardless of the languages, it will be better to use \textsc{CoD} only for those low-resource ones. This can be told visually from Figure \ref{cod_f} that \textsc{CoD} brings better improvements for the bottom figure that the baseline reports lower scores compared to the upper figure with higher baseline scores. The selection can be done with a threshold on the scores, and we observe that for those languages with a baseline score under 20 points in chrF++, \textsc{CoD} brings consistent improvements. We found using our universal list of high-resource auxiliary languages performs well and one can tune the list for specific languages for further improvements.\footnote{We have found putting source and target language at the head of the chain empirically works well via early attempts. We empirically suggest to set the chain length as 5. Further increasing the length can further improve the information, while making the method less cost-effective.}
\par
\paragraph{En-X: COMET Scores} We first obtain 99 languages out of the 200 languages from FLORES-200, which is supported by COMET (this list is obtained by matching the language names to the description in the official COMET repository)\footnote{\url{https://github.com/Unbabel/COMET}} Table \ref{comet} reports COMET scores, which aligns with our previous conclusion and indicates that \textsc{CoD} is effective. The average score of COMET is 0.325 for \textsc{CoD}, which is apparently higher than 0.277 from the baseline. We also found the same conclusion in the remaining 101 languages not perfectly supported by COMET. Since they are not perfectly supported, we do not report those languages here to avoid confusion. 
\subsection{X-En Results}
\paragraph{X-En: ChatGPT}
In addition to the results for translation from English into other languages, we also use our multilingual dictionary for testing translation into English. Table \ref{xen_results} and Table \ref{xen_results2} in the Appendix report the comparison between GPT-3.5-TURBO and \textsc{CoD}. We observe very good improvements in all languages when translating into English. We speculate that the underlying reason is that English is the major language used to pre-train GPT-3.5-TURBO. Dictionaries give hints to the model to produce better translation output by relying on the dictionary vocabulary and predicting the relationship between them. We also found that the translation capacity of ChatGPT can be non-symmetric, e.g., for umb\_Latn, English translation reports a score of 17.41 in chrF++, while translating into English reports a score of 4.64 only.
\begin{table}[t!]
\scriptsize
\centering
    \setlength\tabcolsep{15pt}
    \setlength\extrarowheight{0pt}
\begin{tabular}{l|cc}
\hline
\noalign{\vskip 1mm}  
\textbf{Model} & \textbf{chrF++} & \textbf{BLEU}\\
\noalign{\vskip 1mm}  
\hline
\noalign{\vskip 1mm} 
GPT-3.5  & 35.30  & 12.52\\
Monolingual Dictionary$\dag$ & 31.58 & 10.97\\
Bilingual Dictionary$\ddag$ & 36.37  & 12.63\\
Decomposed Dictionary &31.20 & 8.96\\ 
Few-shot ICL (1) &36.72 & 12.78\\
Few-shot ICL (3) &36.93&12.95\\
\noalign{\vskip 1mm}  
\hline
\noalign{\vskip 1mm} 
\textsc{CoD} (Partially Replaced I) & 37.78 & 13.72\\
\textsc{CoD} (Partially Replaced II) & 37.47 & 13.29\\
\textsc{CoD} (Chain 1)$\dag$& 31.58 & 10.97\\
\textsc{CoD} (Chain 2)$\ddag$& 36.37  & 11.06\\
\textsc{CoD} (Chain 3)& 35.47  & 12.29\\
\textsc{CoD} (Chain 4) &37.90 & 13.90\\
\textsc{CoD} (Chain 5) &\textbf{38.27} & \textbf{13.90}\\
\noalign{\vskip 1mm}  
\hline
\end{tabular}
\caption{\label{ablation_text}
Evaluations of \textsc{CoD} and various baselines on GPT-3.5 averaged from 200 languages. We report on translating from English into other languages. $\dag$,$\ddag$: the models are the same except for their different names.}
\end{table}
\begin{table}[t!]
 \small
\centering
    \setlength\tabcolsep{6pt}
    \setlength\extrarowheight{0pt}
\begin{tabular}{l|ccc}
\hline
\noalign{\vskip 1mm}  
\textbf{Language} & \textbf{BLOOM} & \textbf{CoD} & \textbf{CoD w/o stopwords}\\
\noalign{\vskip 1mm}  
\hline
\noalign{\vskip 1mm} 
srp\_Cyrl & 26.20 & \textbf{39.26} & 38.66\\
tzm\_Ting & 12.55 & 10.93&\textbf{13.12}\\
ckb\_Arab & 7.05 &  \textbf{12.50}&9.83\\
kon\_Tatn & 14.09 & \textbf{17.03} & 14.56\\
smo\_Latn&13.80&15.09&\textbf{16.01}\\
uig\_Arab&11.97&\textbf{14.86}&13.54\\
azb\_Arab&12.42 & \textbf{14.39} & 12.50\\
amh\_Ethi&13.12&\textbf{17.00}&16.82\\
nus\_Latn&13.24&\textbf{14.70}&14.27\\
kac\_Latn&13.25&\textbf{16.28}&14.73\\
\noalign{\vskip 1mm}  
\hline
\end{tabular}
\caption{\label{bloom}
Evaluations in chrF++ of \textsc{CoD} on BLOOM in the direction of translating from other languages into English. We report results on 10 randomly selected low-resource languages on the FLORES-200 full devtest set.
}
\end{table}
\paragraph{X-En: BLOOM} Table \ref{bloom} reports results in chrF++ on BLOOM on 10 randomly selected low-resource languages translating into English. While the improvement is clear (e.g., from 7.05 to 12.50 on ckb\_Arab), the improvement on BLOOM seems less significant than on ChatGPT. One reason could be that we are using a smaller model on BLOOM (7B). This can make the instruction less native to the LLMs as we do not do any instruction tuning or fine-tuning on BLOOM. We leave this to future work for further improvement.
\par
\paragraph{X-En on BLOOM: Save Computations via Removing Stopwords} Table \ref{bloom} truncate stopwords and reduces 4,978 dictionaries from the total of 15,074. The experiments are conducted on 10 randomly selected low-resource languages. The results in chrF++ indicate that such truncation can effectively save about 1/3 of the dictionary prompts, while still maintaining satisfying translation performance. While the original \textsc{CoD} shows better performance in most directions, removing stopwords can even occasionally surpass the original \textsc{CoD}, for example on tzm\_Ting: \textsc{CoD}(10.93), removing stopwords (\textbf{13.12}). We postulate that it is hard for GPTs to translate even those stopwords for low-resource languages.
\par
\subsection{X-Y Results}
\paragraph{X-Y: ChatGPT} Table \ref{xychrf++} compares \textsc{CoD} to GPT-3.5-TURBO on X-Y translations that we randomly select from the 30 languages as experiments with InstructGPT. The languages contain both higher-resourced and lower-resourced ones. \textsc{CoD} brings excellent improvements to 25/30 of the translations, by up to more than 10x improvements (1.33->14.48 in chrF++ scores for srp\_Cyrl->kac\_Latn).
\par
\begin{table}
\centering
    \setlength\tabcolsep{18pt}
    \setlength\extrarowheight{0pt}
\begin{tabular}{l|c}
\hline
\noalign{\vskip 1mm}  
\textbf{Model} & \textbf{FLORES-200}\\
\noalign{\vskip 1mm}  
\hline
\noalign{\vskip 1mm} 
GPT-3.5-TURBO&0.277\\
\noalign{\vskip 1mm}  
\hline
\noalign{\vskip 1mm} 
\textsc{CoD} & \textbf{0.325}\\
\noalign{\vskip 1mm}  
\hline
\end{tabular}
\caption{\label{comet}
Results of COMET scores for 99 supported languages on the FLORES-200 full devtest. We report on translating from English into other languages.
}
\end{table}
\begin{table}
\centering
    \setlength\tabcolsep{15pt}
    \setlength\extrarowheight{0pt}
\begin{tabular}{l|cc}
\hline
\noalign{\vskip 1mm}  
\textbf{Model} & \textbf{X-En} & \textbf{En-X}\\
\noalign{\vskip 1mm}  
\hline
\noalign{\vskip 1mm} 
GPT-3.5-TURBO&44.98&33.22\\
NLLB & 54.77& \textbf{43.39}\\
\noalign{\vskip 1mm}  
\hline
\noalign{\vskip 1mm} 
\textsc{CoD} & \textbf{66.12} & 36.49\\
\noalign{\vskip 1mm}  
\hline
\end{tabular}
\caption{\label{nllb}
Results of \textsc{CoD} (based on GPT-3.5-TURBO) compared to SOTA translator NLLB with chrF++ scores on 200 languages from FLORES-200 full devtest set.
}
\end{table}

\subsection{Comparison to SOTA Translators} 
Table \ref{nllb} reports the translation performance of \textsc{CoD} on both X-En and En-X directions. While NLLB surpasses \textsc{CoD} on EX, we observe that \textsc{CoD} can give a promising performance on X-En and even surpass the SOTA translator NLLB.\footnote{We also found that using perfect English dictionaries on X-En improves \textsc{CoD} from 66.12 to 68.37. This means that our generated dictionaries are of good quality.}
\subsection{Ablation Study}
Table \ref{ablation_text} reports the ablation study using GPT-3.5 that was accessed through the online GUI user interface. More details are in the Appendix \ref{mored}.
\begin{table}
\centering
    \setlength\tabcolsep{6pt}
    \setlength\extrarowheight{0pt}
\begin{tabular}{l|cc}
\hline
\noalign{\vskip 1mm}  
\textbf{Model} & \textbf{chrF++} & \textbf{BLEU}\\
\noalign{\vskip 1mm}  
\hline
\noalign{\vskip 1mm} 
GPT-3.5  & 32.97  & 11.45\\
\citet{2023arXiv230207856G} &  35.60 & 11.58\\
\noalign{\vskip 1mm}  
\hline
\noalign{\vskip 1mm} 
\textsc{CoD} & \textbf{36.30} & \textbf{12.01}\\
\noalign{\vskip 1mm}  
\hline
\end{tabular}
\caption{\label{multilingual_analysis}
Evaluations of \textsc{CoD} and various baselines on GPT-3.5 averaged from 200 languages. We report on translating from English into other languages.
}
\end{table}

\begin{figure*}[t!]
\begin{center}
\vspace{0mm}
\centerline{
%\hspace{-10mm}
\includegraphics[width=14.5cm]{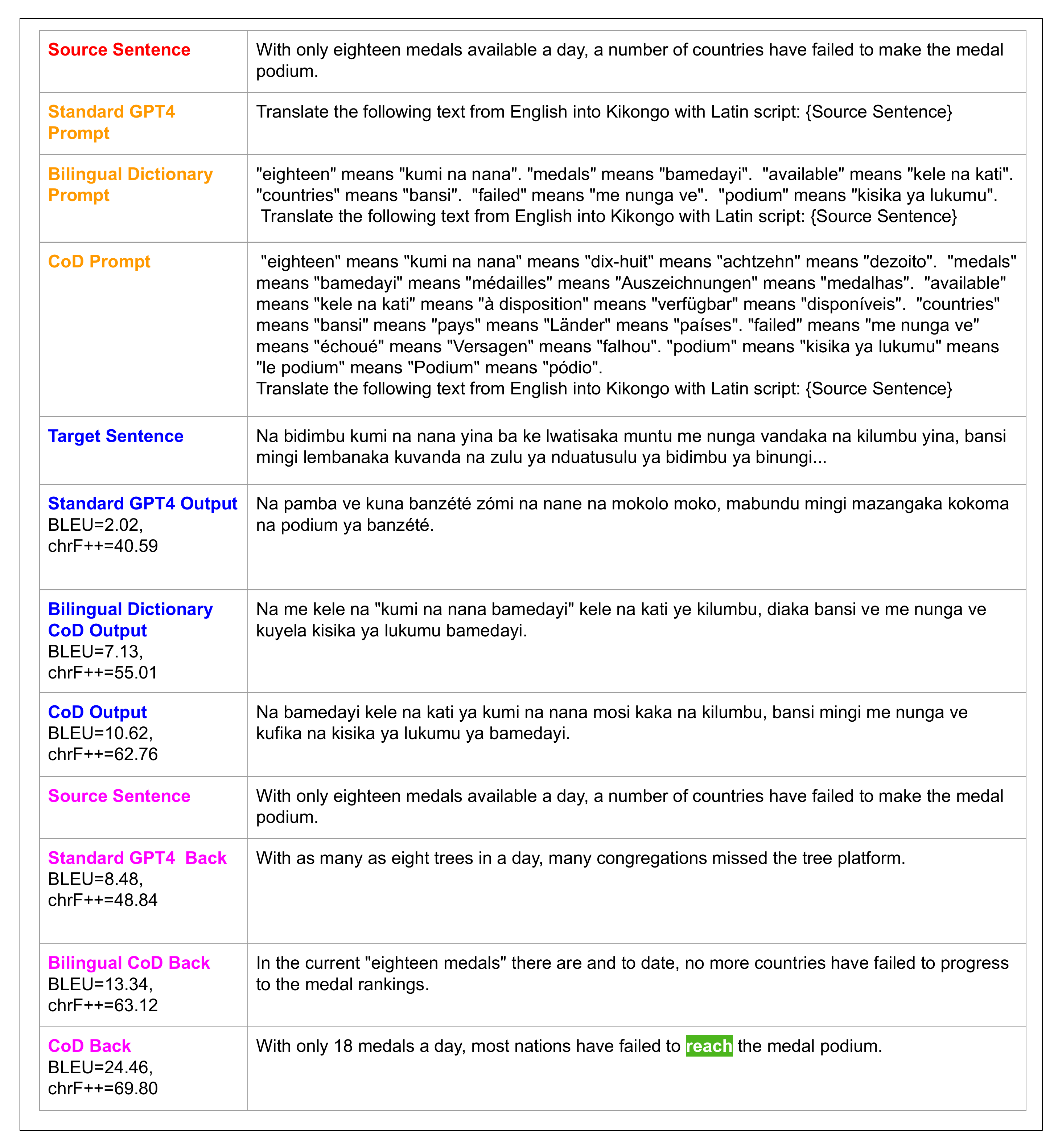}}
    \caption{A case study on translating from English into Kikongo with Latin script using GPT-4 throughout the cases. We evaluate the results on BLEU and chrF++. We highlight in green the words translated wrong by baselines but translated correctly by CoD, even if the words are not presented in the multilingual dictionary chains.}
    \label{fig:case1}
\end{center}
\vspace{-5mm}
\end{figure*}
\paragraph{Multilingual Dictionary} As in Table \ref{ablation_text}, using multilingual dictionaries from \textsc{CoD} instead of using a bilingual dictionary clearly improves the translation performance. Compared to using a bilingual dictionary that brings improvements of 1.07 chrF++ points to GPT-3.5, \textsc{CoD} brings a further improvement of 1.56 points in chrF++. This is more drastic on GPT-3.5-TURBO in Table \ref{multilingual_analysis}, where bilingual dictionary \citep{2023arXiv230207856G} clearly shows lower performance than \textsc{CoD}. In comparison, \textsc{CoD} effectively improves the BLEU score on the baseline from 11.45 to 12.01. Also as in Table \ref{ablation_text}, using a monolingual dictionary with target translation only can be harmful, and we suspect that it can confuse the model as there is no cross-lingual cue in the monolingual dictionary.
\paragraph{Chained Dictionary} Removing chained dictionaries and using non-chained dictionaries that flatten all the dictionaries clearly deteriorates the translation results. We postulate that one reason is that a flattened dictionary introduces repeated source language text as redundant information, which can degrade the results. This claim aligns with the fact in \citet{2023arXiv230200093S} that LLMs can be easily distracted by irrelevant context. Reducing the chaining length (\textsc{CoD} (Chain 1, 2, 3, 4)) also drops the performance. We kindly note that our goal is rather research-oriented. We leave longer chaining to future work, which might yield better performance.
\paragraph{Few-shot In-context Learning (ICL)} Retrieving few-shot demonstrations for in-context learning instead of \textsc{CoD} for languages in FLORES-200 brings minor improvement. We postulate that the reason is the difficulty in understanding low-resource languages, and therefore the retrieved demonstrations are still not very useful to the desired translation. While increasing the number of demonstrations in the prompt can further boost the performance, the results are still not very promising, below \textsc{CoD}.
\paragraph{Selection of Auxiliary Languages}
Partially replacing the auxiliary language (\textsc{CoD} (Partially Replaced I, II)) to arbitrary other languages (for example, Arabic (arb\_Arab) instead of high-resource German (deu\_Latn)) drops the performance.\footnote{We also found that using other languages that are similar to the target language, such as the languages written in the same script, can lead to an obvious drop in performance. We suspect that putting a similar language to the target language tends to produce those languages in the output. However, using high-resource language in Latin script as the auxiliary language does not suffer from such a problem.} We should use more high-resource languages in the chain for better performance. We suspect that such high-resource languages yield stronger cross-lingual hints to be used for the translations.
\subsection{Case Study}
Figure \ref{fig:case1} presents a case study demonstrating the powerfulness of \textsc{CoD}. The baseline output from GPT4 is almost lost about which topics are discussed in the sentence. Using a bilingual dictionary is useful, but the bilingual baseline is still lost about the detailed semantics. In comparison, \textsc{CoD} successfully provides a high-quality translation, scoring the best in BLEU and chrF++. We also highlight in green where the translation is successfully elicited by \textsc{CoD}, even if the words are not provided in the multilingual dictionary. We hypothesise that \textsc{CoD} provides richer context to the LLMs to translate relevant words in the source sentences, even if they are not directly presented by \textsc{CoD}. Figure \ref{fig:case99} and Figure \ref{fig:case999} demonstrate cases that show a similar phenomenon, and they are available in the Appendix, at the end of this paper.
\section{Related Work}
\paragraph{Neural Machine Translation via Prompting Language Models}
Limited research has been conducted on effective methods for prompting large language models in machine translation. The majority of existing research has concentrated on evaluating the translation capabilities of large language models, utilizing uncomplicated prompts such as `Translate to {language\_name}: {text}' \citep{NEURIPS2020_1457c0d6,2021arXiv211210668L,2022arXiv221105100W,2022arXiv220501068Z}. Various prompt formats have been explored by the scholars \citep{10.1145/3411763.3451760,2023arXiv230304048W}, whereas \citet{2022arXiv220211822G} have examined the potential use of prompts for regulating the formality or specific dialect of the output. Furthermore, \citet{2022arXiv221202437A} and \citet{2022arXiv221109102V} have focused on identifying appropriate in-context examples to improve machine translation quality with LLMs.
\paragraph{Lexical-based Neural Machine Translation}
Our research is connected to the concept of lexical restrictions in MT, which can be categorized into either hard constraints \citep{hokamp-liu-2017-lexically,post-vilar-2018-fast} or soft constraints \citep{2019arXiv190409107S,dinu-etal-2019-training,10.5555/3491440.3491936}.
\par
Also, several works have explored the use of dictionaries in supervised MT. \citet{2016arXiv161007272Z} improves NMT with a bilingual dictionary that includes less common or unseen words present in the bilingual training data. \citet{arthur-etal-2016-incorporating} enhances the translation of infrequent words by supplementing the system with discrete translation lexicons and utilizing the attention vector to select the pertinent lexical probabilities. \citet{10.1145/3377713.3377801} uses a dictionary to generate synthetic parallel data to better train the NMT models. A previous work uses bilingual dictionaries to improve MT \citep{2023arXiv230207856G}.
\par
\textsc{CoD} is one of the first applications of applying dictionaries on Machine Translation on LLMs. Note that this paper focuses on proving the effectiveness of applying a dictionary to LLMs rather than providing an actual dictionary to be used.
\section{Conclusions}
\textsc{CoD} is a novel framework that uses chained multilingual dictionaries when prompting large language models (LLMs) for MNMT. We evaluate ChatGPT, InstructGPT, and BLOOM on the FLORES-200 dataset for MNMT. We found that ChatGPT and InstructGPT still have room for improvement in translating many language pairs.  \textsc{CoD} elicits large gains by up to 13x chrF++ points for MNMT (3.08 to 42.63 for English to Serbian written in Cyrillic script) on FLORES-200 full devtest set. We also verified the necessity of the chained multilingual dictionaries, and we found that both of them are quite important to \textsc{CoD}. \textsc{CoD} also outperforms few-shot demonstrations which struggle to retrieve relevant demonstrations for low-resource settings. \textsc{CoD} can even surpass the strong SOTA NLLB translator in translation. Extensive case studies demonstrate that \textsc{CoD} elicits translation even if the words are not directly presented by \textsc{CoD}.
\section*{Limitations}
This paper presents an analysis of 200 languages only. However, there are more than thousands of languages around the world.
\par 
Although \textsc{CoD} can lead to a very slight degradation in translation performance for a small subset of languages, our experiments have shown that the impact is typically insignificant and can be probably simply due to randomness. Therefore, the practical usage of \textsc{CoD} remains unaffected.
\par 
While \textsc{CoD} brings by up to 1.8x inference time as found in our implementation, the inference time for actual LLM APIs can be down to milliseconds, so this is realistic to apply \textsc{CoD} to real products. 
\par 
While \textsc{CoD} brings by up to 3x prompt length, many LLMs support very long input lengths, for example, 32K for GPT4. So this is realistic to apply \textsc{CoD} to real products. One can also save the tokens by prompting rare words only with \textsc{CoD}.
\par
This work also does not directly compare to those ones that require fine-tuning on LLMs \citep{jiao-etal-2023-parrot} which requires error-guided data. Nevertheless, \textsc{CoD} is easy to use and does not require additional data. It is comparatively easy to curate good-quality dictionaries with off-the-shelf tools.
\par 
We also consider and focus on the task of Machine Translation, as it is one of the most fundamental NLG tasks.
\section*{Ethical Statement}
We honour and support the EMNLP Code of Ethics. There is no ethical issue known to us. A well-known and widely used LLM is used in our work, which is subjected to generating offensive context. However, the above-mentioned issues are widely known to commonly exist for LLMs. Any content generated does not reflect the view of the authors.
\begin{comment}
End-to-end pre-trained dialogue generators using large-scale text corpus are also employed. The generators can be subjected to offensive contexts and demographic or historical biases hidden in the training data. Despite the fact that the model releasers have attempted their efforts to reduce such contexts in their training data, the model retains the potential to trigger offensive replies and might express agreement towards offensive or unethical contexts. Since current state-of-the-art end-to-end pre-trained dialogue generators or pre-trained language models are mostly trained on large corpus or conversations that naturally occur, the above-mentioned issues are widely known to commonly exist for these models. Any contents generated do not reflect the view of the authors.
\end{comment}
\bibliography{anthology,custom}
\bibliographystyle{acl_natbib}
\appendix
\begin{table*}[t!]
\tiny
\centering
    \setlength\tabcolsep{6pt}
\setlength\aboverulesep{0pt}\setlength\belowrulesep{0pt}
\setcellgapes{0pt}\makegapedcells
\begin{tabular}{l|cc|l|cc|l|cc|l|cc|l|cc}
\hline
\noalign{\vskip 1mm}  
\textbf{Language} & \textbf{GPT} & \textbf{CoD} & \textbf{Language} & \textbf{GPT} & \textbf{CoD} & \textbf{Language} & \textbf{GPT} & \textbf{CoD} & \textbf{Language} & \textbf{GPT} & \textbf{CoD} & \textbf{Language} & \textbf{GPT} & \textbf{CoD}\\
\noalign{\vskip 1mm}  
\hline
\noalign{\vskip 1mm}  
ace\_Arab&10.96&\hlc[aqua]{\textbf{12.87}}&ace\_Latn&24.38&\hlc[aqua]{\textbf{30.94}}&acm\_Arab&\hlc[aqua]{\textbf{40.16}}&38.37&acq\_Arab&\hlc[aqua]{\textbf{43.46}}&40.02&aeb\_Arab&\hlc[aqua]{\textbf{38.16}}&36.19\\afr\_Latn&\hlc[aqua]{\textbf{65.25}}&64.71&ajp\_Arab&\hlc[aqua]{\textbf{43.38}}&42.47&aka\_Latn&22.01&\hlc[aqua]{\textbf{25.69}}&als\_Latn&\hlc[aqua]{\textbf{52.60}}&51.64&amh\_Ethi&10.05&\hlc[aqua]{\textbf{19.93}}\\apc\_Arab&\hlc[aqua]{\textbf{42.60}}&41.24&arb\_Arab&\hlc[aqua]{\textbf{49.85}}&49.08&ars\_Arab&\hlc[aqua]{\textbf{46.68}}&45.13&ary\_Arab&\hlc[aqua]{\textbf{33.53}}&32.04&arz\_Arab&\hlc[aqua]{\textbf{39.25}}&38.77\\asm\_Beng&19.83&\hlc[aqua]{\textbf{27.11}}&ast\_Latn&\hlc[aqua]{\textbf{52.81}}&52.52&awa\_Deva&32.16&\hlc[aqua]{\textbf{32.47}}&ayr\_Latn&21.05&\hlc[aqua]{\textbf{25.76}}&azb\_Arab&2.96&\hlc[aqua]{\textbf{18.11}}\\azj\_Latn&34.17&\hlc[aqua]{\textbf{36.65}}&bak\_Cyrl&20.15&\hlc[aqua]{\textbf{31.90}}&bam\_Latn&17.43&\hlc[aqua]{\textbf{23.02}}&ban\_Latn&31.68&\hlc[aqua]{\textbf{35.63}}&bel\_Cyrl&33.90&\hlc[aqua]{\textbf{35.00}}\\bem\_Latn&22.63&\hlc[aqua]{\textbf{27.46}}&ben\_Beng&35.29&\hlc[aqua]{\textbf{38.08}}&bho\_Deva&27.98&\hlc[aqua]{\textbf{29.43}}&bjn\_Arab&12.06&\hlc[aqua]{\textbf{13.35}}&bjn\_Latn&32.88&\hlc[aqua]{\textbf{36.87}}\\bod\_Tibt&20.70&\hlc[aqua]{\textbf{24.37}}&bos\_Latn&\hlc[aqua]{\textbf{56.10}}&55.38&bug\_Latn&17.62&\hlc[aqua]{\textbf{26.56}}&bul\_Cyrl&\hlc[aqua]{\textbf{58.23}}&57.73&cat\_Latn&\hlc[aqua]{\textbf{63.29}}&62.19\\
\hline
\end{tabular}
\caption{\label{enx_results}
Comparison between GPT-3.5-TURBO and \textsc{CoD}. Results in chrF++ for MT on the FLORES-200 dataset. The best results are bolded and highlighted. We report on translating from English into the languages.}
\end{table*}
\begin{table*}[t!]
\tiny
\centering
    \setlength\tabcolsep{6pt}
    \setlength\extrarowheight{0pt}
\begin{tabular}{l|cc|l|cc|l|cc|l|cc|l|cc}
\hline
\noalign{\vskip 1mm}  
\textbf{Language} & \textbf{GPT} & \textbf{CoD} & \textbf{Language} & \textbf{GPT} & \textbf{CoD} & \textbf{Language} & \textbf{GPT} & \textbf{CoD} & \textbf{Language} & \textbf{GPT} & \textbf{CoD} & \textbf{Language} & \textbf{GPT} & \textbf{CoD}\\
\noalign{\vskip 1mm}  
\hline
\noalign{\vskip 1mm}  
ace\_Arab&3.43&\hlc[aqua]{\textbf{44.72}}&ace\_Latn&10.20&\hlc[aqua]{\textbf{57.40}}&acm\_Arab&28.57&\hlc[aqua]{\textbf{59.86}}&acq\_Arab&29.49&\hlc[aqua]{\textbf{61.15}}&aeb\_Arab&24.18&\hlc[aqua]{\textbf{54.56}}\\afr\_Latn&53.42&\hlc[aqua]{\textbf{73.29}}&ajp\_Arab&33.14&\hlc[aqua]{\textbf{63.78}}&aka\_Latn&7.33&\hlc[aqua]{\textbf{46.52}}&als\_Latn&33.69&\hlc[aqua]{\textbf{63.26}}&amh\_Ethi&3.84&\hlc[aqua]{\textbf{50.16}}\\apc\_Arab&30.26&\hlc[aqua]{\textbf{61.71}}&arb\_Arab&33.75&\hlc[aqua]{\textbf{63.22}}&ars\_Arab&31.83&\hlc[aqua]{\textbf{62.53}}&ary\_Arab&21.72&\hlc[aqua]{\textbf{53.29}}&arz\_Arab&25.74&\hlc[aqua]{\textbf{55.55}}\\asm\_Beng&12.10&\hlc[aqua]{\textbf{52.60}}&ast\_Latn&36.38&\hlc[aqua]{\textbf{60.59}}&awa\_Deva&19.87&\hlc[aqua]{\textbf{54.48}}&ayr\_Latn&4.44&\hlc[aqua]{\textbf{42.24}}&azb\_Arab&8.61&\hlc[aqua]{\textbf{49.86}}\\azj\_Latn&17.48&\hlc[aqua]{\textbf{46.86}}&bak\_Cyrl&8.87&\hlc[aqua]{\textbf{47.07}}&bam\_Latn&4.95&\hlc[aqua]{\textbf{48.20}}&ban\_Latn&17.35&\hlc[aqua]{\textbf{58.12}}&bel\_Cyrl&17.16&\hlc[aqua]{\textbf{41.73}}\\bem\_Latn&7.58&\hlc[aqua]{\textbf{47.98}}&ben\_Beng&20.56&\hlc[aqua]{\textbf{59.26}}&bho\_Deva&15.54&\hlc[aqua]{\textbf{49.91}}&bjn\_Arab&4.06&\hlc[aqua]{\textbf{41.10}}&bjn\_Latn&19.08&\hlc[aqua]{\textbf{60.84}}\\bod\_Tibt&2.18&\hlc[aqua]{\textbf{43.64}}&bos\_Latn&37.91&\hlc[aqua]{\textbf{63.31}}&bug\_Latn&7.41&\hlc[aqua]{\textbf{48.21}}&bul\_Cyrl&35.93&\hlc[aqua]{\textbf{63.12}}&cat\_Latn&42.26&\hlc[aqua]{\textbf{65.33}}\\
\hline
\end{tabular}
\caption{\label{xen_results}
Comparison of \textsc{CoD} against GPT-3.5-TURBO. Results in chrF++ for MT on the FLORES-200 dataset. The best results are bolded and highlighted. We report on translating from the languages into English.}
\end{table*}

\begin{table*}[t!]
\tiny
\centering
    \setlength\tabcolsep{9pt}
    \setlength\extrarowheight{1pt}
\begin{tabular}{l|cc|l|cc|l|cc|l|cc}
\hline
\noalign{\vskip 1mm}  
\textbf{Language} & \textbf{GPT} & \textbf{CoD} & \textbf{Language} & \textbf{GPT} & \textbf{CoD} & \textbf{Language} & \textbf{GPT} & \textbf{CoD} & \textbf{Language} & \textbf{GPT} & \textbf{CoD}\\
\noalign{\vskip 1mm}  
\hline
\noalign{\vskip 1mm}  
ace\_Arab&-0.41&\hlc[aqua]{\textbf{-0.22}}&ace\_Latn&-0.97&\hlc[aqua]{\textbf{-0.52}}&acm\_Arab&\hlc[aqua]{\textbf{0.72}}&0.67&acq\_Arab&\hlc[aqua]{\textbf{0.65}}&0.58\\aeb\_Arab&\hlc[aqua]{\textbf{0.70}}&0.63&afr\_Latn&\hlc[aqua]{\textbf{0.48}}&0.37&ajp\_Arab&-0.34&\hlc[aqua]{\textbf{-0.13}}&aka\_Latn&-0.72&\hlc[aqua]{\textbf{-0.71}}\\als\_Latn&-0.48&\hlc[aqua]{\textbf{-0.32}}&amh\_Ethi&\hlc[aqua]{\textbf{0.66}}&0.59&apc\_Arab&0.28&\hlc[aqua]{\textbf{0.40}}&arb\_Arab&0.33&\hlc[aqua]{\textbf{0.38}}\\ars\_Arab&\hlc[aqua]{\textbf{0.71}}&0.65&ary\_Arab&\hlc[aqua]{\textbf{0.85}}&0.81&arz\_Arab&\hlc[aqua]{\textbf{0.42}}&0.32&asm\_Beng&0.01&\hlc[aqua]{\textbf{0.21}}\\ast\_Latn&-0.03&\hlc[aqua]{\textbf{0.05}}&awa\_Deva&-0.24&\hlc[aqua]{\textbf{-0.23}}&ayr\_Latn&-0.89&\hlc[aqua]{\textbf{-0.74}}&azb\_Arab&-0.03&\hlc[aqua]{\textbf{0.22}}\\azj\_Latn&\hlc[aqua]{\textbf{0.62}}&0.54&bak\_Cyrl&-0.46&\hlc[aqua]{\textbf{-0.08}}&bam\_Latn&-1.46&\hlc[aqua]{\textbf{0.68}}&ban\_Latn&-0.60&\hlc[aqua]{\textbf{-0.51}}\\bel\_Cyrl&\hlc[aqua]{\textbf{0.84}}&0.80&bem\_Latn&-0.67&\hlc[aqua]{\textbf{-0.49}}&ben\_Beng&-0.13&\hlc[aqua]{\textbf{0.02}}&bho\_Deva&-0.46&\hlc[aqua]{\textbf{-0.31}}\\
\hline
\end{tabular}
\caption{\label{enx_results3}
Comparison between GPT-3.5-TURBO and \textsc{CoD}. Results in COMET for MT on the FLORES-200 dataset. The best results are bolded and highlighted. We report on translating from English into the languages.}
\end{table*}
\begin{table*}[t!]
\tiny
\centering
    \setlength\tabcolsep{1pt}
    \setlength\extrarowheight{0pt}
\begin{tabular}{c|cc|c|cc|c|cc|c|cc|c|cc}
\hline
\noalign{\vskip 1mm}  
\textbf{Direction} & \textbf{GPT} & \textbf{CoD} & \textbf{Direction} & \textbf{GPT} & \textbf{CoD} & \textbf{Direction} & \textbf{GPT} & \textbf{CoD} & \textbf{Direction} & \textbf{GPT} & \textbf{CoD} & \textbf{Direction} & \textbf{GPT} & \textbf{CoD}\\
\noalign{\vskip 1mm}  
\hline
\noalign{\vskip 1mm}  
amh\_Ethi->lao\_Laoo&15.43&\hlc[aqua]{\textbf{16.40}}&azb\_Arab->tsn\_Latn&20.68&\hlc[aqua]{\textbf{24.81}}&bak\_Cyrl->amh\_Ethi&7.68&\hlc[aqua]{\textbf{10.72}}&bug\_Latn->tgk\_Cyrl&15.41&\hlc[aqua]{\textbf{16.16}}&ckb\_Arab->tzm\_Tfng&\hlc[aqua]{\textbf{8.68}}&7.72\\hau\_Latn->kac\_Latn&4.51&\hlc[aqua]{\textbf{11.69}}&hye\_Armn->tsn\_Latn&22.56&\hlc[aqua]{\textbf{24.00}}&ibo\_Latn->hye\_Armn&\hlc[aqua]{\textbf{16.74}}&16.47&kac\_Latn->srp\_Cyrl&6.93&\hlc[aqua]{\textbf{11.55}}&kbp\_Latn->shn\_Mymr&4.73&\hlc[aqua]{\textbf{6.99}}\\kir\_Cyrl->bug\_Latn&10.17&\hlc[aqua]{\textbf{14.10}}&kon\_Latn->srp\_Cyrl&\hlc[aqua]{\textbf{5.01}}&3.72&lao\_Laoo->snd\_Arab&\hlc[aqua]{\textbf{12.61}}&7.80&lin\_Latn->zul\_Latn&21.35&\hlc[aqua]{\textbf{23.00}}&nso\_Latn->bug\_Latn&10.65&\hlc[aqua]{\textbf{16.40}}\\nya\_Latn->sag\_Latn&15.57&\hlc[aqua]{\textbf{18.13}}&plt\_Latn->nso\_Latn&23.60&\hlc[aqua]{\textbf{28.42}}&sag\_Latn->lin\_Latn&21.70&\hlc[aqua]{\textbf{24.23}}&shn\_Mymr->amh\_Ethi&4.39&\hlc[aqua]{\textbf{5.92}}&smo\_Latn->lao\_Laoo&19.36&\hlc[aqua]{\textbf{19.84}}\\snd\_Arab->bug\_Latn&8.26&\hlc[aqua]{\textbf{15.68}}&sot\_Latn->amh\_Ethi&8.86&\hlc[aqua]{\textbf{10.83}}&srp\_Cyrl->kac\_Latn&1.33&\hlc[aqua]{\textbf{14.48}}&tat\_Cyrl->hye\_Armn&22.22&\hlc[aqua]{\textbf{23.51}}&tgk\_Cyrl->amh\_Ethi&8.82&\hlc[aqua]{\textbf{11.30}}\\tsn\_Latn->plt\_Latn&23.99&\hlc[aqua]{\textbf{25.14}}&tso\_Latn->sot\_Latn&\hlc[aqua]{\textbf{25.90}}&25.77&tzm\_Tfng->amh\_Ethi&3.42&\hlc[aqua]{\textbf{3.43}}&uig\_Arab->tgk\_Cyrl&14.94&\hlc[aqua]{\textbf{17.74}}&zul\_Latn->amh\_Ethi&8.75&\hlc[aqua]{\textbf{11.19}}\\
\hline
\end{tabular}
\caption{\label{xychrf++}
Comparison of \textsc{CoD} against GPT-3.5-TURBO. Results in chrF++ for MT on the FLORES-200 dataset. The best results are bolded and highlighted. We report on translating from X into Y.}
\end{table*}
\begin{table*}[t!]
\tiny
\centering
    \setlength\tabcolsep{9pt}
    \setlength\extrarowheight{2.7pt}
\setcellgapes{0pt}\makegapedcells
\begin{tabular}{l|cc|l|cc|l|cc|l|cc}
\hline
\noalign{\vskip 1mm}  
\textbf{Language} & \textbf{GPT} & \textbf{CoD} & \textbf{Language} & \textbf{GPT} & \textbf{CoD} & \textbf{Language} & \textbf{GPT} & \textbf{CoD} & \textbf{Language} & \textbf{GPT} & \textbf{CoD} \\
\noalign{\vskip 1mm}  
\hline
\noalign{\vskip 1mm}  
ace\_Arab&1.88&\hlc[aqua]{\textbf{1.94}}&ace\_Latn&5.19&\hlc[aqua]{\textbf{5.95}}&acm\_Arab&\hlc[aqua]{\textbf{11.31}}&10.77&acq\_Arab&\hlc[aqua]{\textbf{13.81}}&13.56\\aeb\_Arab&\hlc[aqua]{\textbf{10.61}}&9.37&afr\_Latn&\hlc[aqua]{\textbf{36.89}}&36.22&ajp\_Arab&13.07&\hlc[aqua]{\textbf{13.10}}&aka\_Latn&4.25&\hlc[aqua]{\textbf{4.54}}\\als\_Latn&\hlc[aqua]{\textbf{23.99}}&23.11&amh\_Ethi&1.86&\hlc[aqua]{\textbf{3.38}}&apc\_Arab&\hlc[aqua]{\textbf{12.24}}&11.27&arb\_Arab&\hlc[aqua]{\textbf{20.11}}&18.97\\ars\_Arab&\hlc[aqua]{\textbf{16.87}}&16.50&ary\_Arab&\hlc[aqua]{\textbf{7.84}}&7.29&arz\_Arab&\hlc[aqua]{\textbf{11.20}}&10.73&asm\_Beng&3.02&\hlc[aqua]{\textbf{4.75}}\\ast\_Latn&\hlc[aqua]{\textbf{22.41}}&21.97&awa\_Deva&7.33&\hlc[aqua]{\textbf{7.36}}&ayr\_Latn&\hlc[aqua]{\textbf{3.69}}&3.48&azb\_Arab&2.23&\hlc[aqua]{\textbf{2.50}}\\azj\_Latn&8.44&\hlc[aqua]{\textbf{8.86}}&bak\_Cyrl&3.48&\hlc[aqua]{\textbf{5.41}}&bam\_Latn&3.05&\hlc[aqua]{\textbf{3.58}}&ban\_Latn&7.59&\hlc[aqua]{\textbf{8.65}}\\bel\_Cyrl&8.50&\hlc[aqua]{\textbf{8.73}}&bem\_Latn&4.40&\hlc[aqua]{\textbf{5.12}}&ben\_Beng&8.66&\hlc[aqua]{\textbf{9.37}}&bho\_Deva&6.26&\hlc[aqua]{\textbf{6.83}}\\
\hline
\end{tabular}
\caption{\label{bleu}
Comparison of \textsc{CoD} against GPT-3.5-TURBO. Results in BLEU for MT on the FLORES-200 dataset. The best results are bolded and highlighted. We report on translating from English into the languages.}
\end{table*}
 
\begin{table*}[t!]
\tiny
\centering
    \setlength\tabcolsep{6pt}
    \setlength\extrarowheight{0pt}
\begin{tabular}{l|cc|l|cc|l|cc|l|cc|l|cc}
\hline
\noalign{\vskip 1mm}  
\textbf{Language} & \textbf{GPT} & \textbf{CoD} & \textbf{Language} & \textbf{GPT} & \textbf{CoD} & \textbf{Language} & \textbf{GPT} & \textbf{CoD} & \textbf{Language} & \textbf{GPT} & \textbf{CoD} & \textbf{Language} & \textbf{GPT} & \textbf{CoD}\\
\noalign{\vskip 1mm}  
\hline
\noalign{\vskip 1mm}  
srp\_Cyrl&10.19&\hlc[aqua]{\textbf{22.18}}&kac\_Latn&9.27&\hlc[aqua]{\textbf{20.38}}&ckb\_Arab&18.73&\hlc[aqua]{\textbf{32.59}}&azb\_Arab&21.40&\hlc[aqua]{\textbf{25.44}}&tzm\_Tfng&16.87&\hlc[aqua]{\textbf{31.00}}\\kon\_Latn&34.07&\hlc[aqua]{\textbf{40.00}}&tat\_Cyrl&26.62&\hlc[aqua]{\textbf{36.36}}&nso\_Latn&19.73&\hlc[aqua]{\textbf{30.46}}&sag\_Latn&13.05&\hlc[aqua]{\textbf{29.22}}&bak\_Cyrl&11.15&\hlc[aqua]{\textbf{20.55}}\\shn\_Mymr&21.73&\hlc[aqua]{\textbf{31.61}}&lin\_Latn&21.79&\hlc[aqua]{\textbf{33.80}}&uig\_Arab&23.57&\hlc[aqua]{\textbf{32.35}}&hau\_Latn&25.01&\hlc[aqua]{\textbf{34.22}}&ibo\_Latn&23.59&\hlc[aqua]{\textbf{32.24}}\\amh\_Ethi&23.27&\hlc[aqua]{\textbf{32.52}}&zul\_Latn&27.89&\hlc[aqua]{\textbf{36.19}}&bug\_Latn&16.46&\hlc[aqua]{\textbf{27.75}}&lao\_Laoo&5.66&\hlc[aqua]{\textbf{22.29}}&tso\_Latn&28.26&\hlc[aqua]{\textbf{37.79}}\\kbp\_Latn&16.22&\hlc[aqua]{\textbf{28.17}}&tsn\_Latn&23.51&\hlc[aqua]{\textbf{32.12}}&smo\_Latn&5.28&\hlc[aqua]{\textbf{46.61}}&snd\_Arab&19.90&\hlc[aqua]{\textbf{33.19}}&hye\_Armn&22.44&\hlc[aqua]{\textbf{33.29}}\\nya\_Latn&23.49&\hlc[aqua]{\textbf{31.70}}&sot\_Latn&25.32&\hlc[aqua]{\textbf{33.68}}&tgk\_Cyrl&2.02&\hlc[aqua]{\textbf{18.49}}&plt\_Latn&8.15&\hlc[aqua]{\textbf{29.03}}&kir\_Cyrl&25.33&\hlc[aqua]{\textbf{35.22}}\\
\noalign{\vskip 1mm}  
\hline
\end{tabular}
\caption{\label{instructchrf++}
Comparison of \textsc{CoD} against TEXT-DAVINCI-003. Results in chrF++ for MT on the FLORES-200 dataset. The best results are bolded and highlighted. We report on translating from the languages into English.}
\end{table*}
\begin{table*}[t!]
\small
\centering
    \setlength\tabcolsep{4.5pt}
    \setlength\extrarowheight{0pt}
\begin{tabular}{l|cc|l|cc|l|cc|l|cc}
\hline
\noalign{\vskip 1mm}  
\textbf{Language} & \textbf{GPT} & \textbf{CoD} & \textbf{Language} & \textbf{GPT} & \textbf{CoD} & \textbf{Language} & \textbf{GPT} & \textbf{CoD} & \textbf{Language} & \textbf{GPT} & \textbf{CoD}\\
\noalign{\vskip 1mm}  
\hline
\noalign{\vskip 1mm}  
ace\_Arab&3.35&\hlc[aqua]{\textbf{45.43}}&ace\_Latn&10.12&\hlc[aqua]{\textbf{56.56}}&acm\_Arab&27.78&\hlc[aqua]{\textbf{59.66}}&acq\_Arab&29.45&\hlc[aqua]{\textbf{61.01}}\\aeb\_Arab&24.38&\hlc[aqua]{\textbf{54.72}}&afr\_Latn&53.62&\hlc[aqua]{\textbf{72.57}}&ajp\_Arab&33.45&\hlc[aqua]{\textbf{62.98}}&aka\_Latn&7.87&\hlc[aqua]{\textbf{46.86}}\\als\_Latn&33.73&\hlc[aqua]{\textbf{62.52}}&amh\_Ethi&4.10&\hlc[aqua]{\textbf{51.04}}&apc\_Arab&29.70&\hlc[aqua]{\textbf{61.44}}&arb\_Arab&33.30&\hlc[aqua]{\textbf{62.87}}\\ars\_Arab&32.31&\hlc[aqua]{\textbf{61.93}}&ary\_Arab&21.67&\hlc[aqua]{\textbf{53.13}}&arz\_Arab&25.54&\hlc[aqua]{\textbf{55.65}}&asm\_Beng&12.09&\hlc[aqua]{\textbf{53.20}}\\ast\_Latn&36.22&\hlc[aqua]{\textbf{60.15}}&awa\_Deva&19.51&\hlc[aqua]{\textbf{53.83}}&ayr\_Latn&4.49&\hlc[aqua]{\textbf{43.40}}&azb\_Arab&8.37&\hlc[aqua]{\textbf{49.53}}\\azj\_Latn&16.96&\hlc[aqua]{\textbf{46.19}}&bak\_Cyrl&8.63&\hlc[aqua]{\textbf{47.04}}&bam\_Latn&4.85&\hlc[aqua]{\textbf{48.18}}&ban\_Latn&17.36&\hlc[aqua]{\textbf{59.05}}\\bel\_Cyrl&17.05&\hlc[aqua]{\textbf{42.43}}&bem\_Latn&7.84&\hlc[aqua]{\textbf{47.90}}&ben\_Beng&20.66&\hlc[aqua]{\textbf{58.67}}&bho\_Deva&14.90&\hlc[aqua]{\textbf{50.09}}\\
\hline
\end{tabular}
\caption{\label{xen_results2}
Comparison of \textsc{CoD} against GPT-3.5-TURBO. Results in chrF++ for MT on the FLORES-200 dataset. The best results are bolded and highlighted. We report on translating from the languages into English.}
\end{table*}

\begin{table*}[t!]
\tiny
\centering
    \setlength\tabcolsep{6pt}
    \setlength\extrarowheight{0pt}
\begin{tabular}{l|cc|l|cc|l|cc|l|cc|l|cc}
\hline
\noalign{\vskip 1mm}  
\textbf{Language} & \textbf{GPT} & \textbf{CoD} & \textbf{Language} & \textbf{GPT} & \textbf{CoD} & \textbf{Language} & \textbf{GPT} & \textbf{CoD} & \textbf{Language} & \textbf{GPT} & \textbf{CoD} & \textbf{Language} & \textbf{GPT} & \textbf{CoD}\\
\noalign{\vskip 1mm}  
\hline
\noalign{\vskip 1mm}  
srp\_Cyrl&2.08&\hlc[aqua]{\textbf{4.84}}&kac\_Latn&2.22&\hlc[aqua]{\textbf{2.73}}&ckb\_Arab&3.24&\hlc[aqua]{\textbf{5.42}}&azb\_Arab&3.92&\hlc[aqua]{\textbf{4.43}}&tzm\_Tfng&2.55&\hlc[aqua]{\textbf{4.54}}\\kon\_Latn&8.33&\hlc[aqua]{\textbf{11.27}}&tat\_Cyrl&4.70&\hlc[aqua]{\textbf{6.96}}&nso\_Latn&3.86&\hlc[aqua]{\textbf{7.12}}&sag\_Latn&1.78&\hlc[aqua]{\textbf{4.75}}&bak\_Cyrl&2.32&\hlc[aqua]{\textbf{3.19}}\\shn\_Mymr&3.49&\hlc[aqua]{\textbf{5.15}}&lin\_Latn&3.54&\hlc[aqua]{\textbf{7.22}}&uig\_Arab&7.80&\hlc[aqua]{\textbf{8.46}}&hau\_Latn&4.43&\hlc[aqua]{\textbf{7.28}}&ibo\_Latn&4.33&\hlc[aqua]{\textbf{7.50}}\\amh\_Ethi&4.45&\hlc[aqua]{\textbf{5.75}}&zul\_Latn&4.73&\hlc[aqua]{\textbf{7.11}}&bug\_Latn&2.58&\hlc[aqua]{\textbf{5.01}}&lao\_Laoo&1.30&\hlc[aqua]{\textbf{1.38}}&tso\_Latn&5.83&\hlc[aqua]{\textbf{10.96}}\\kbp\_Latn&2.41&\hlc[aqua]{\textbf{5.51}}&tsn\_Latn&4.15&\hlc[aqua]{\textbf{6.76}}&smo\_Latn&3.83&\hlc[aqua]{\textbf{17.43}}&snd\_Arab&3.55&\hlc[aqua]{\textbf{5.84}}&hye\_Armn&3.88&\hlc[aqua]{\textbf{6.69}}\\nya\_Latn&3.75&\hlc[aqua]{\textbf{6.57}}&sot\_Latn&4.28&\hlc[aqua]{\textbf{6.72}}&tgk\_Cyrl&2.03&\hlc[aqua]{\textbf{3.00}}&plt\_Latn&2.35&\hlc[aqua]{\textbf{4.38}}&kir\_Cyrl&4.10&\hlc[aqua]{\textbf{6.18}}\\
\noalign{\vskip 1mm}  
\hline
\end{tabular}
\caption{\label{instructbleu}
Comparison of \textsc{CoD} against TEXT-DAVINCI-003. Results in BLEU for MT on the FLORES-200 dataset. The best results are bolded and highlighted. We report on translating from the languages into English.}
\end{table*}
\section{More Experimental Details}
\label{mored}
For the ablation study with GPT-3.5, We manually tested 800 instances from the FLORES-200 dataset that covers all the languages. For the ablation study with GPT-3.5-TURBO, we report the full devset evaluations.

\section{Creating the Dictionary}
Other tools such as FastAlign \citep{dyer-etal-2013-simple} can also be used for word alignment in creating dictionaries with bilingual corpora.
\section{InstructGPT}
\label{inst}
Table  \ref{instructchrf++} and Table \ref{instructbleu} compare \textsc{CoD} against TEXT-DAVINCI-003 on 30 languages that we found \textsc{CoD} works well on ChatGPT from FLORES-200 full devtest set. The results indicate that \textsc{CoD} improves all of them on InstructGPT as well, with an average boost of 12.02 in chrF++ (from 18.99 to 31.01) and  2.61 in BLEU (from 3.73 to 6.34).
\begin{figure*}[t!]
\begin{center}
\vspace{0mm}
\centerline{
%\hspace{-10mm}
\includegraphics[width=13cm]{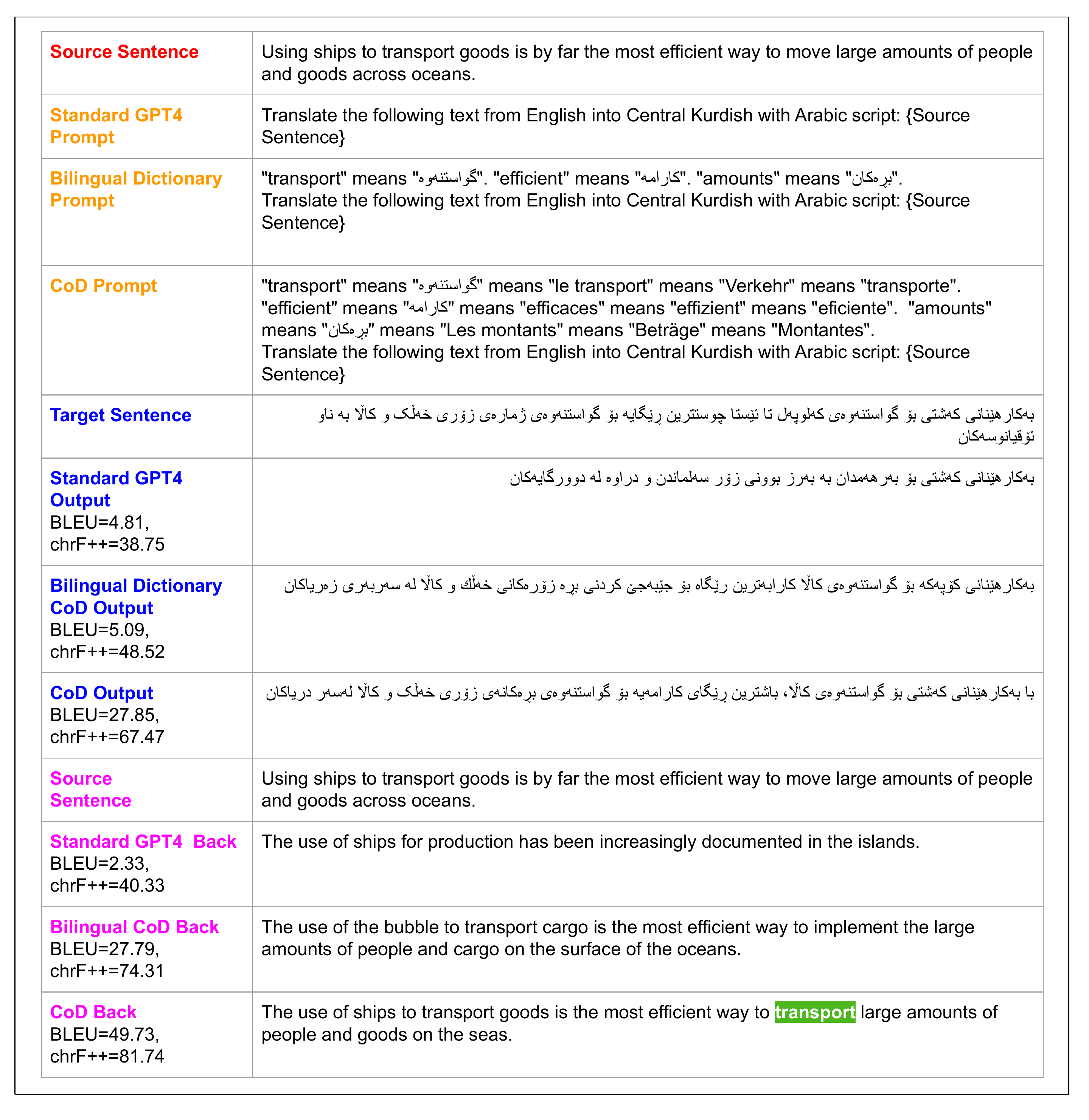}}
    \caption{A case study on translating from English into Central Kurdish with Latin script using GPT-4 throughout the cases. We evaluate the results on BLEU and chrF++. We highlight in green the words translated wrong by baselines but translated correctly by CoD, even if the words are not presented in the multilingual dictionary chains.}
    \label{fig:case99}
\end{center}
\vspace{-5mm}
\end{figure*}

\begin{figure*}[t!]
\begin{center}
\vspace{0mm}
\centerline{
%\hspace{-10mm}
\includegraphics[width=13cm]{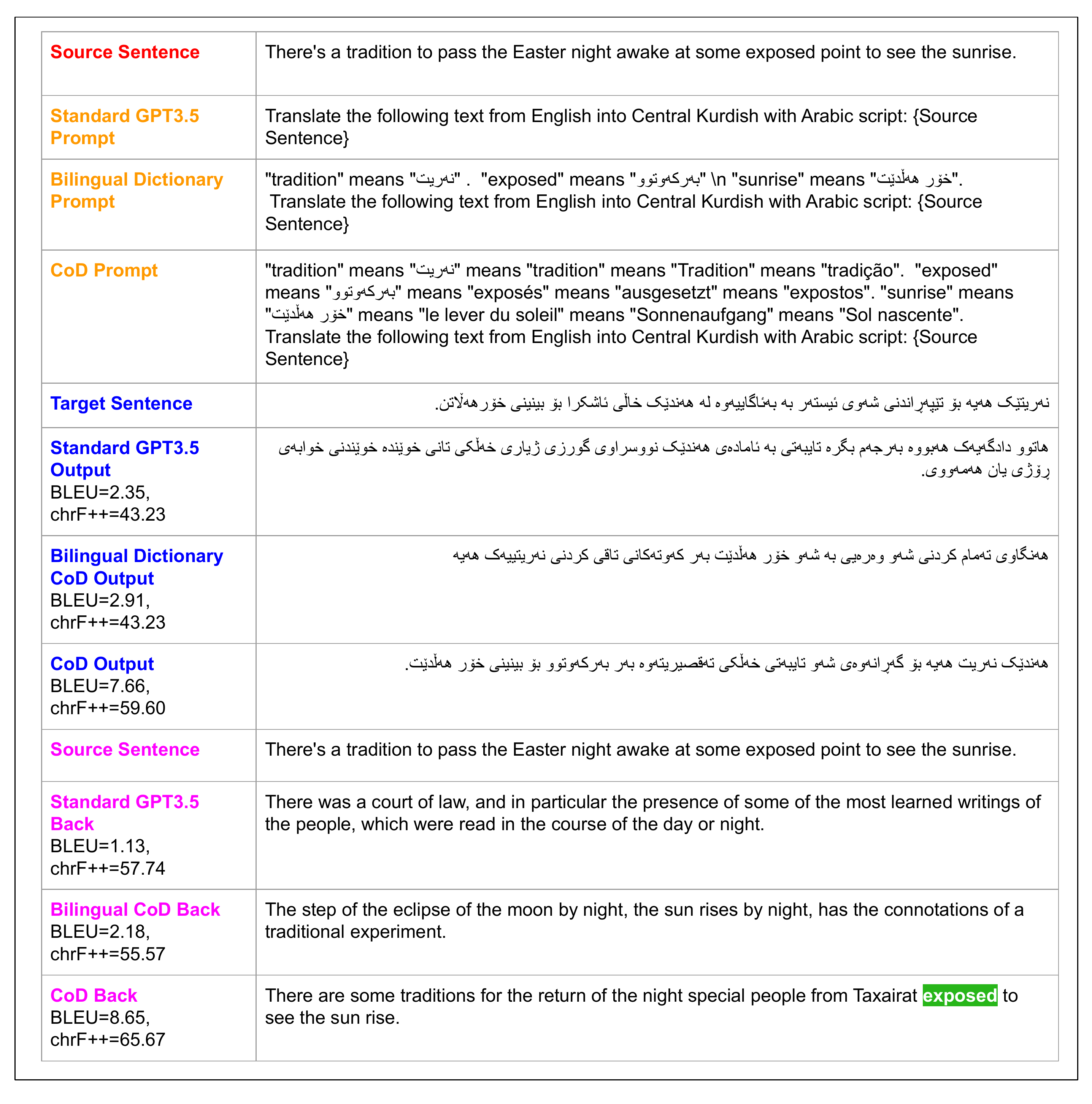}}
    \caption{A case study on translating from English into Central Kurdish with Latin script using GPT-3.5 throughout the cases. We evaluate the results on BLEU and chrF++. We highlight in green the words translated wrong by baselines but translated correctly by CoD, even if the words are not presented in the multilingual dictionary chains.}
    \label{fig:case999}
\end{center}
\vspace{-5mm}
\end{figure*}

\end{document}